%% file: main.tex
\documentclass{bmvc2k}

\usepackage{multirow}
\usepackage{pifont}
\usepackage{cleveref}
\usepackage{subcaption}
\usepackage{geometry}
\usepackage{parskip} 
\usepackage{wrapfig}
\usepackage[final]{pdfpages}

\newcommand{\cmark}{\ding{52}}%
\newcommand{\xmark}{\ding{56}}%
\title{REACT: Real-time Efficiency and Accuracy Compromise for Tradeoffs in Scene Graph Generation}

\addauthor{Maëlic Neau}{maelic.neau@umu.se}{1,3,4 $\dagger$}
\addauthor{Paulo E. Santos}{paulo.santos@priorianalytica.com}{2,3}
\addauthor{Anne-Gwenn Bosser}{bosser@enib.fr}{4}
\addauthor{Cédric Buche}{cedric.buche@cnrs.fr}{5,6}
\addauthor{Akihiro Sugimoto}{sugimoto@nii.ac.jp}{7}

\addinstitution{
 Umeå University, Universitetstorget 4, Umeå, Sweden
}
\addinstitution{
 PrioriAnalytica, 74 Pirie St, Adelaide, Australia
}
\addinstitution{
 Flinders University, Sturt Rd, Bedford Park, Australia
}
\addinstitution{
 Bretagne INP - ENIB, 945 Av. du Technopôle, Plouzané, France
}
\addinstitution{
 CNRS IRL 2010 CROSSING, 11 Victoria Drive, Adelaide, Australia
}
\addinstitution{
 IMT Atlantique, 655 Av. du Technopôle, Plouzané, France
}
\addinstitution{
 National Institute of Informatics, 2-12 Hitotsubashi, Tokyo, Japan
}

\runninghead{NEAU, SANTOS, BOSSER, BUCHE, SUGIMOTO}{REACT}

\newcommand\blfootnote[1]{
    \begingroup
    \renewcommand\thefootnote{}\footnote{#1}
    \addtocounter{footnote}{-1}
    \endgroup
}

\begin{document}

\maketitle
\blfootnote{$\dagger$ Corresponding authors.}
\input{sec/0_abstract}    
\input{sec/1_intro}
\input{sec/2_related_work}
\input{sec/3_bottlenecks}
\input{sec/4_real_time_SGG}
\input{sec/5_experiments}
\input{sec/6_ablation_studies}
\input{sec/8_conclusion}

\bibliography{SGG_biblio}

\includepdf[pages=-]{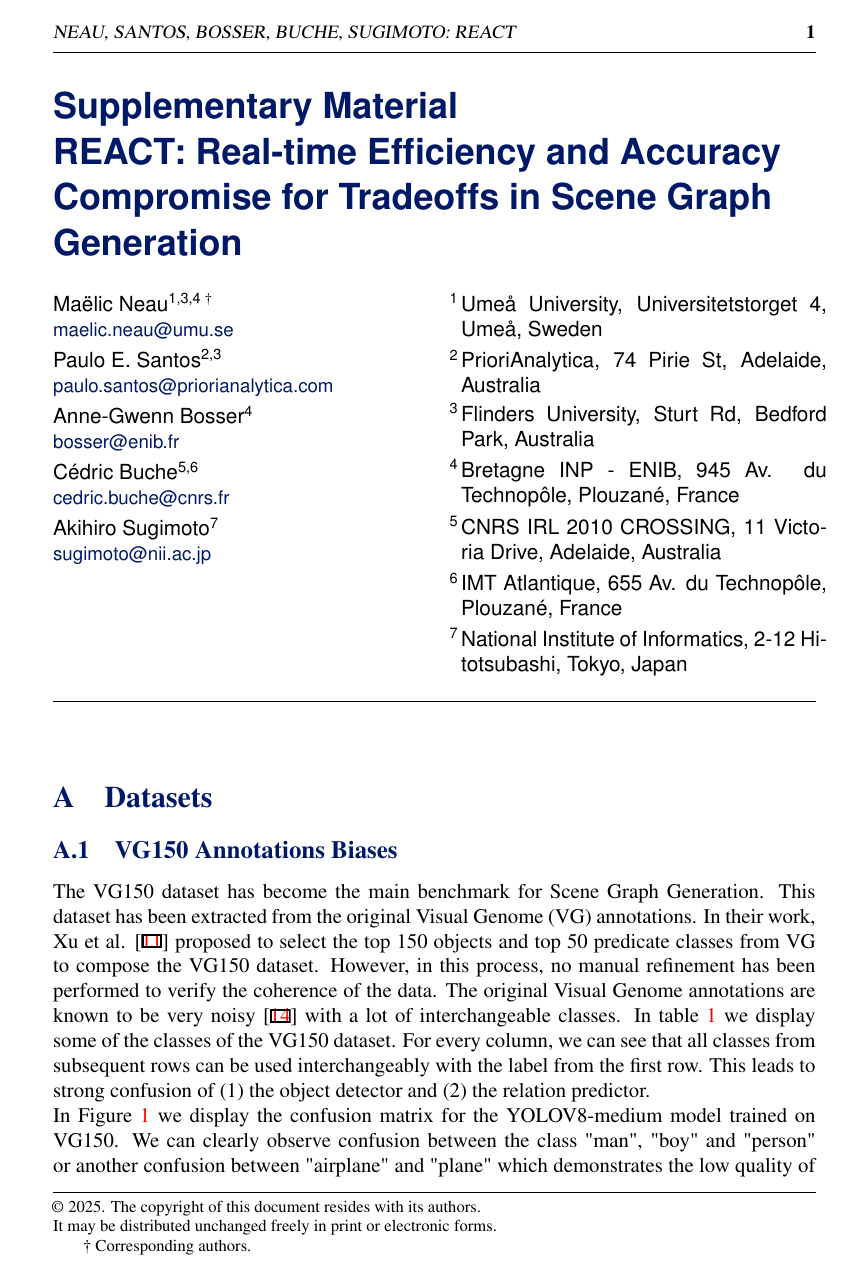}

\end{document}

%% file: sec/0_abstract.tex
\begin{abstract}

Scene Graph Generation (SGG) is a task that encodes visual relationships between objects in images as graph structures. SGG shows significant promise as a foundational component for downstream tasks, such as reasoning for embodied agents. To enable real-time applications, SGG must address the trade-off between performance and inference speed. However, current methods tend to focus on one of the following: (1) improving relation prediction accuracy, (2) enhancing object detection accuracy, or (3) reducing latency, without aiming to balance all three objectives simultaneously.
To address this limitation, we propose the Real-time Efficiency and Accuracy Compromise for Tradeoffs in Scene Graph Generation (REACT) architecture, which achieves the highest inference speed among existing SGG models, improving object detection accuracy without sacrificing relation prediction performance. Compared to state-of-the-art approaches, REACT is 2.7 times faster and improves object detection accuracy by 58\%. Furthermore, our proposal significantly reduces model size, with an average of 5.5x fewer parameters. The code is available at \url{https://github.com/Maelic/SGG-Benchmark}.

\end{abstract}

\newpage

%% file: sec/1_intro.tex
\begin{wrapfigure}{r}{0.55\textwidth}    
    \includegraphics[width=\linewidth]{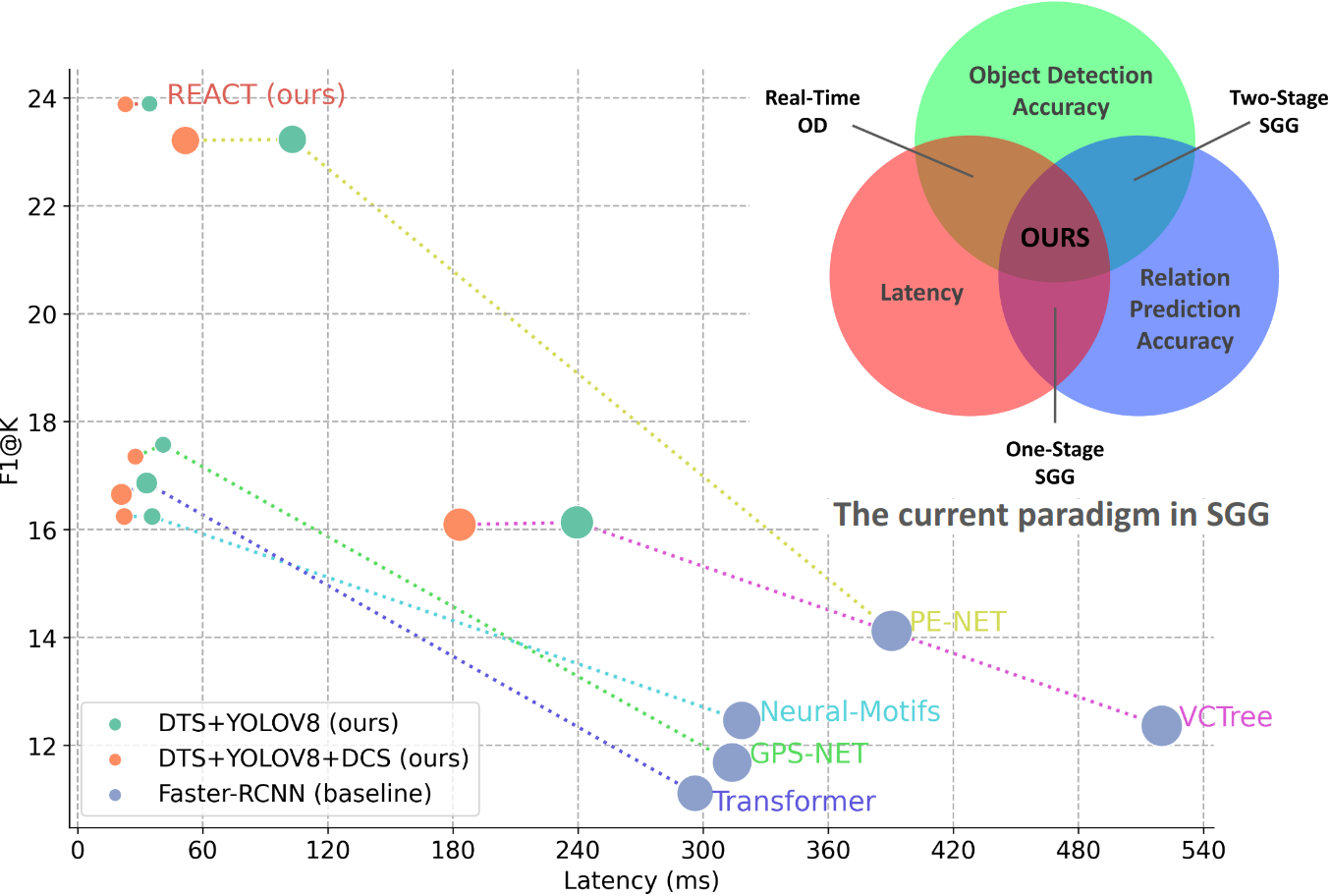}
    \caption{Performance comparison between our approach and previous work, in terms of performance (F1@K metric) and Latency (ms) on the PSG dataset \cite{yang_panoptic_2022}. Points are scaled by the number of parameters (smaller is better).}
    \label{fig:comparison_intro}
\end{wrapfigure}
\section{Introduction}
\label{sec:intro}
Scene Graph Generation (SGG) is the task of generating a structured graph representation from visual inputs. Given an image, visual regions corresponding to objects of interest are extracted, and relations between these objects are predicted in the form of \textit{<subject, predicate, object>} triplets.
The set of these triplets forms a directed acyclic graph which can be interpreted as a structured representation of the scene.
This task has garnered increasing attention recently thanks to its utility in various downstream applications, such as Visual Question Answering (VQA) \cite{hudsonGQANewDataset2019,damodaran2021understanding,liangGraghVQALanguageGuidedGraph2021}, Image Captioning \cite{yang2022reformer,wangRoleSceneGraphs2019,nguyenDefenseSceneGraphs2021a}, or Reasoning of an embodied agent \cite{gadre2022continuous,li2022embodied,neau2023defense}.
However, we observe a significant imbalance between the progress in the SGG task itself and its adoption in downstream tasks \cite{changComprehensiveSurveyScene2023a}. 
One of the reasons is the lack of low-cost approaches that could power applications with real-time constraints. Because SGG is a task that combines Object Detection (OD) and Relation Prediction (RelPred), a good trade-off between accuracy in these two parameters is also required. 
In this work, we explore the bottlenecks in SGG to find an optimal combination of OD accuracy, RelPred accuracy, and latency. SGG methods can be either Two-Stage (TS) or One-Stage (OS) approaches. The former uses a combination of an object detector and a relation predictor to model objects and relations in a sequence. The latter uses a single-stage pipeline to infer relations and object proposals directly from the image features. As a result, one-stage approaches are often more efficient for real-time processing \cite{imEGTRExtractingGraph2024}. 
However, they fall short in accuracy in the OD task \cite{li_sgtr_2022,im2024egtr}. 
\qquad In this work, we investigate the bottlenecks of the two-stage architecture in terms of OD accuracy and latency. The structural dependencies between blocks of the architecture are reviewed as a Decoupled Two-Stage SGG (DTS-SGG), which allows 
maintaining the performance of the pre-trained detector, providing better control of the information flow (i.e., detected object proposals) between the two stages of the architecture. 
This approach reduces the overall latency, as the complexity of relation prediction increases exponentially with the number of input objects. To select an optimal amount of proposals at inference time, we define a new algorithm, the Dynamic Candidate Selection (DCS). DCS uses evaluation results to find the best trade-off between latency and accuracy at test time.
Thanks to the DTS architecture, we can switch the dependency on a Faster-RCNN model for any real-time object detectors, such as one of the YOLO family (YOLOV8/9/10 \cite{Jocher_Ultralytics_YOLO_2023,wang2024yolov9,wang2024yolov10}).
Recent state-of-the-art SGG models such as PE-NET \cite{zheng2023prototype} use a combination of different feature sources, such as union pair features \cite{zellers_neural_2018} or textual features of the subject and object classes \cite{luVisualRelationshipDetection2016a}. This results in a significant amount of processing to gather and refine each type of feature altogether. By removing the dependency on sub-optimal features, we can reduce the parameter count of the model and lower the latency. By coupling this strategy to the DTS architecture, we propose a new model for SGG, the Real-Time Efficiency and Accuracy Compromise for Tradeoffs in SGG (\textbf{REACT}). 
Thus, the main contributions of this work are: (i) a new architecture for SGG, the Decoupled Two-Stage (DTS); 
(ii) a new inference method, Dynamic Candidate Selection (DCS), which finds an optimal number of proposals to select for the relation prediction stage, to further improve latency without compromising accuracy; (iii) a new model, REACT, that extends previous work and our DTS architecture, reducing the latency while enhancing performance on standard metrics, see \Cref{fig:comparison_intro}. 

%% file: sec/2_related_work.tex
\section{Related Work}
\qquad {\bf Two-stage SGG approaches} typically use Faster-RCNN with a VGG-16 \cite{xu_scene_2017, yang_graph_2018} or ResNeXt-101 backbone for object detection and feature extraction, as seen in Neural-Motifs \cite{zellers_neural_2018}, VCTree \cite{tang_learning_2019}, Transformer \cite{tang_unbiased_2020}, GPS-Net \cite{lin_gps-net_2020}, and PE-NET \cite{zheng2023prototype}. In these approaches, object classes probabilities given by Faster-RCNN are refined during relationship learning, via a dedicated classifier \cite{zellers_neural_2018, lin_gps-net_2020} or by aggregating features from neighboring regions \cite{yang_graph_2018}. However, this design can hinder the performance of SGG models on the task of Object Detection. In our work, we choose to effectively \textit{Decoupled} the Object Detector and Relationship Predictor by removing the dedicated classifier and keeping original class probabilities.
In addition, Faster-RCNN’s large backbone limits real-time use. Recent work \cite{jinFastContextualScene2023a} proposed a real-time SGG method using contextual cues, relying on YOLOv5 \cite{Jocher_YOLOv5_by_Ultralytics_2020} bounding boxes to learn correlations between box coordinates and predicates—excluding visual features. However, the method's reliance on spatial features will neglect fine-grained relations such as the difference between $eating$ or $drinking$. Another approach \cite{jinIndependentRelationshipDetection2023} uses a Relation-aware YOLO (RYOLO) to predict relation coordinates via new anchors, matched to nearby objects. Yet, lacking subject-object context leads to poor RelPred performance (3.1\% mR@100) \cite{zellers_neural_2018}. In contrast, we propose to use the features extracted by a YOLO model to generate representations for the subject and object nodes, making the first stage of the architecture more efficient without compromising performance on the second stage. 

\qquad {\bf One-stage SGG approaches} jointly learn object and relation representations. Sparse-RCNN \cite{teng2022structured} uses triplet queries to produce objects and relations, decoded via cascade-RCNN after CNN feature extraction. RelTr \cite{congRelTRRelationTransformer2022} and SGTR \cite{liSGTREndtoendScene2024}, built on DETR \cite{carion2020end}, generate sparse relation sets, unlike dense two-stage predictions. 
More recently,  EGTR \cite{imEGTRExtractingGraph2024} builds upon Deformable-DETR for efficient object detection, paired with graph-based relation matching. RelTr and EGTR report lower latency than two-stage models like Motifs \cite{zellers_neural_2018} and VCTree \cite{tang_learning_2019}, though these comparisons often rely on Faster-RCNN. As highlighted in recent work \cite{liSGTREndtoendScene2024}, learning both the object and relation representations at the same time can make the performance of one-stage models in OD sub-optimal in comparison to two-stage approaches.
We argue that replacing Faster-RCNN with YOLO in two-stage pipelines can achieve both lower latency than two-stage approaches and higher OD accuracy than one-stage approaches, as we shall see in this paper.

\begin{figure}[t]
    \centering
    \begin{subfigure}{0.49\textwidth}
        \centering
        \includegraphics[width=\columnwidth]{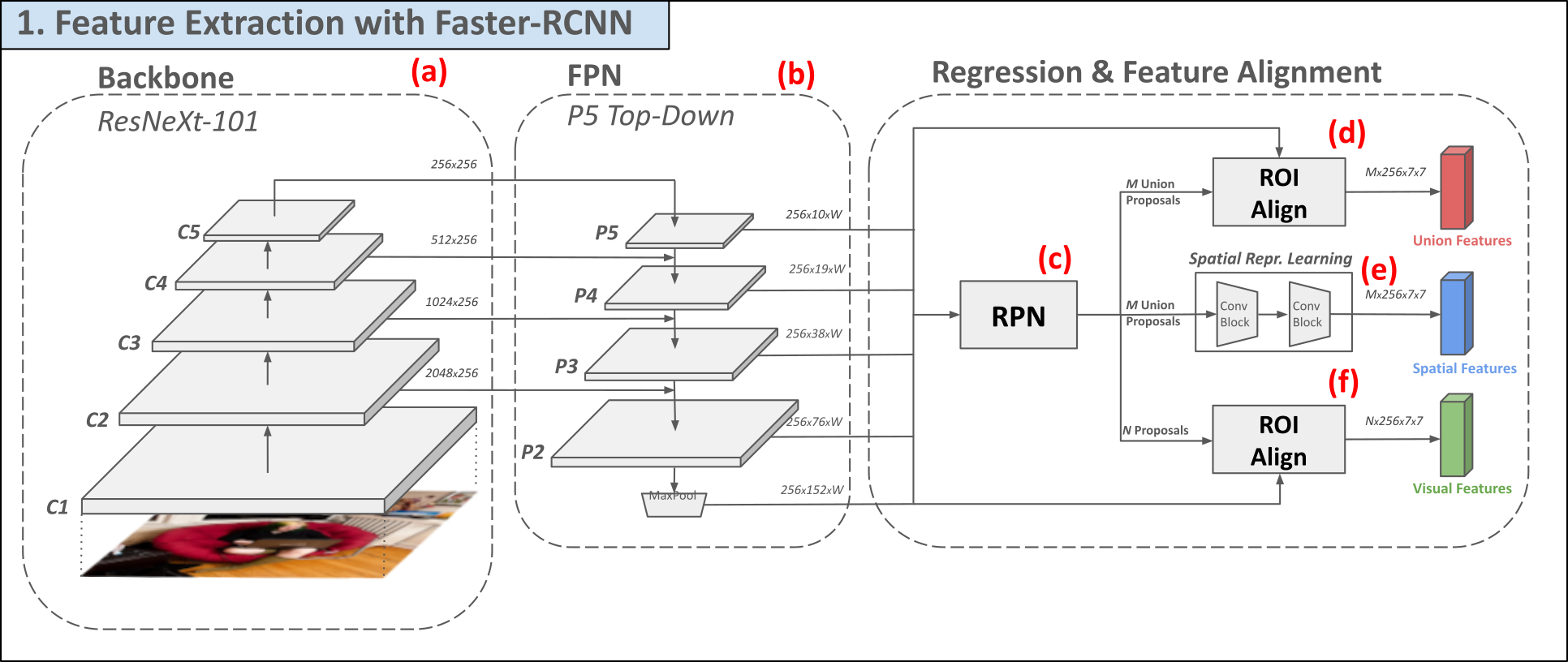}    
        \caption{Stage 1 - Two-Stage Faster-RCNN}
        \label{fig:stage1_penet}
    \end{subfigure}
    \begin{subfigure}{0.49\textwidth}
        \centering
        \includegraphics[width=\columnwidth]{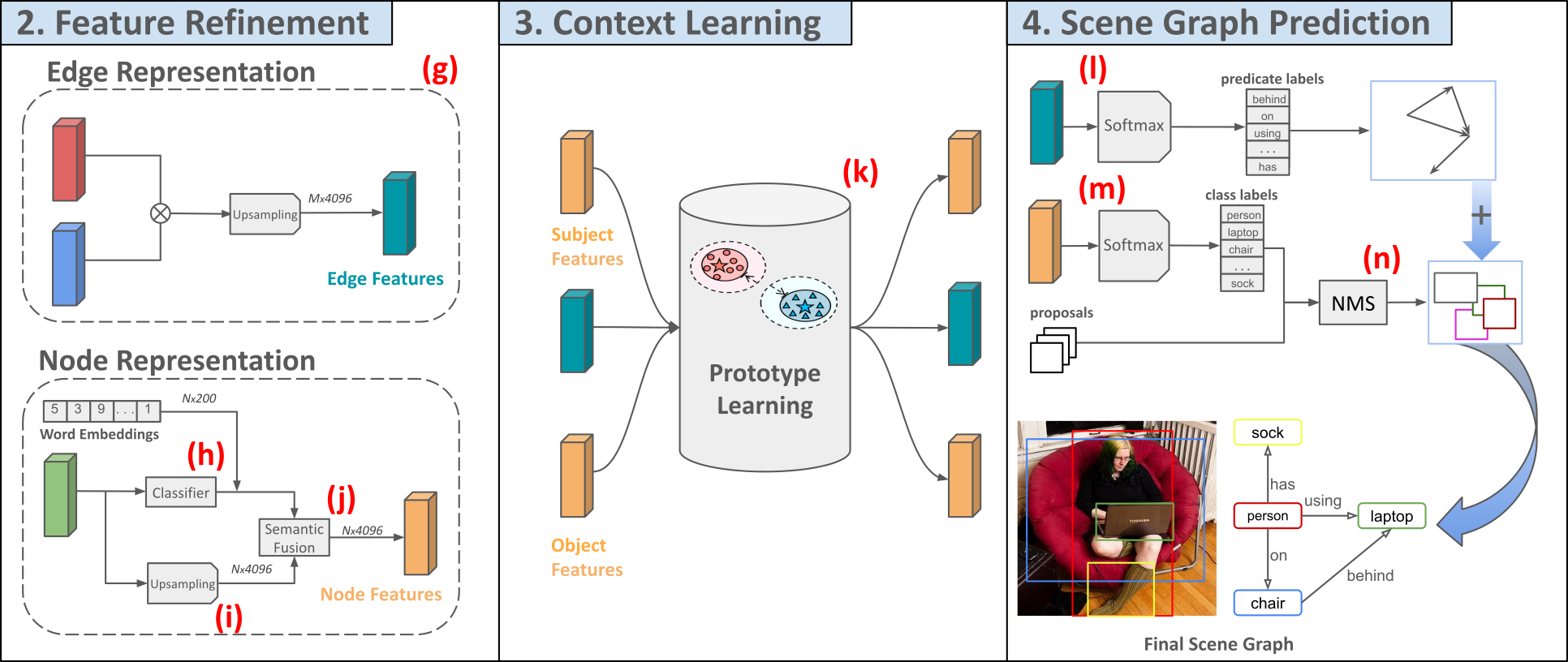}
        \caption{Stage 2 - Prototype Learning (PE-NET)}
        \label{fig:stage2_penet}
    \end{subfigure}
    \begin{subfigure}{0.49\textwidth}
        \centering
        \includegraphics[width=\columnwidth]{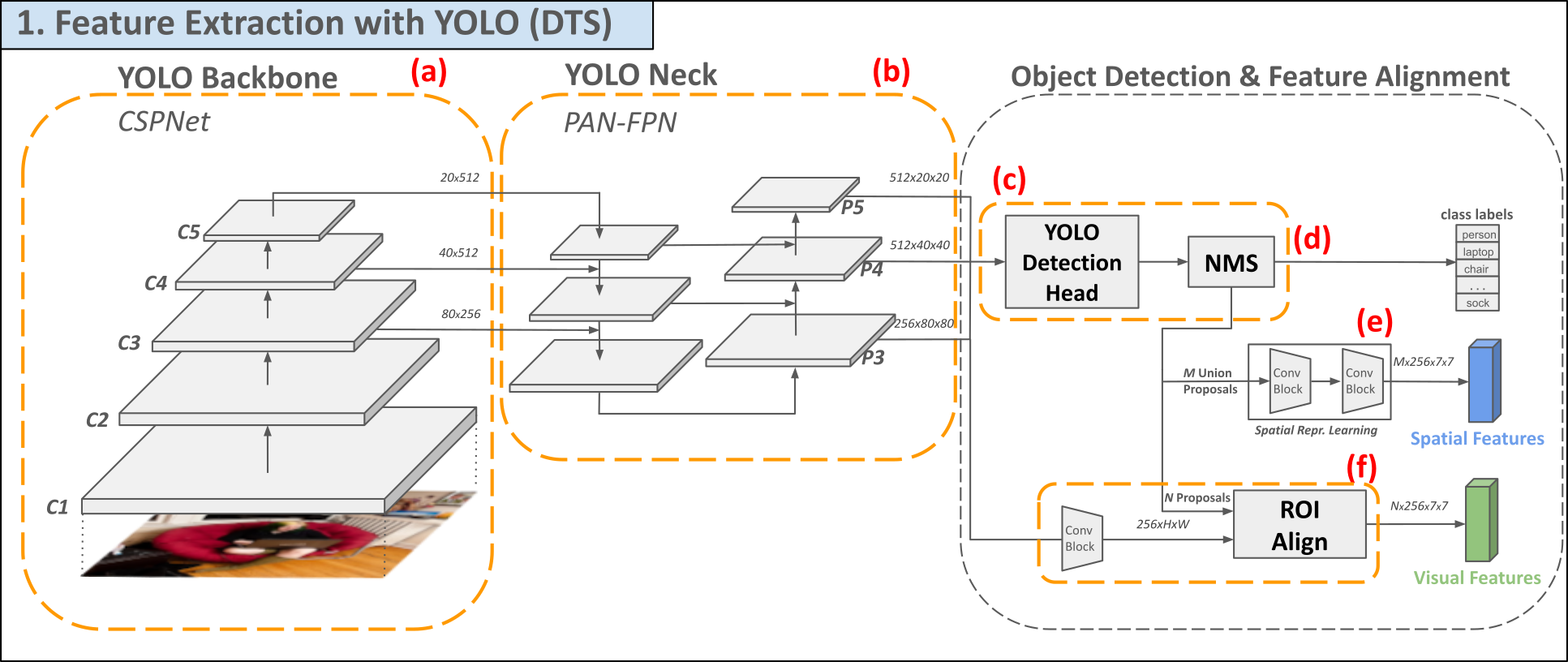}    
        \caption{Stage 1 - DTS YOLO (ours)}
        \label{fig:stage1_react}
    \end{subfigure}
    \begin{subfigure}{0.49\textwidth}
        \centering
        \includegraphics[width=\columnwidth]{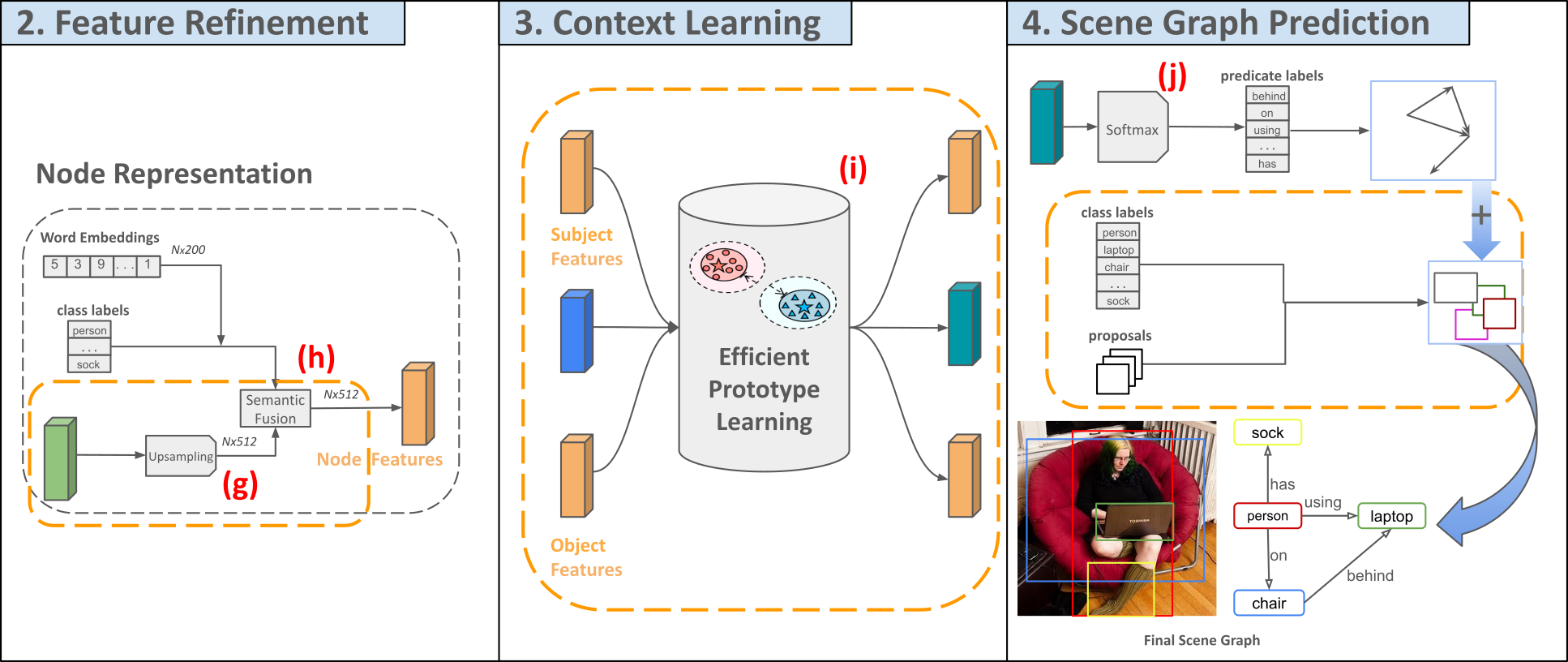}    
        \caption{Stage 2 - Efficient Prototype Learning (ours)}
        \label{fig:stage2_react}
    \end{subfigure}
    \caption[Pipeline of a typical SGG model]{Top: the PE-NET architecture \cite{zheng2023prototype}. Bottom: our proposed REACT model with (left) Decoupled Two-Stage (DTS) + YOLO for efficient object detection and feature extraction ; (right) our proposed Efficient Prototype Learning model. [\textcolor{orange}{\textbf{- - -}}] represent modified or added components, $\bigotimes$ denotes element-wise concatenation. }
    \label{fig:sgg_pipeline}
\end{figure}

\newpage

%% file: sec/3_bottlenecks.tex
\section{Investigating Bottlenecks in SGG}\label{sec:bottlenecks}

\Cref{fig:sgg_pipeline} (top row) is a representation of the PE-NET model \cite{zheng2023prototype}: a state-of-the-art two-stage SGG approach. The feature extraction (stage 1 \Cref{fig:stage1_penet}) as well as the graph prediction (module 4, \Cref{fig:stage2_penet}) are common to all two-stage architectures \cite{tang_learning_2019,tang_unbiased_2020,dong_stacked_2022,li_bipartite_2021,zellers_neural_2018,lin_gps-net_2020}.
In Stage 1, visual features are extracted by a ResNeXt-101 backbone (\Cref{fig:stage1_penet} \textbf{(a)}), and a Feature Pyramid Network (FPN) \cite{lin2017feature} is used to refine the features at different scales (\Cref{fig:stage1_penet} \textbf{(b)}). Then, a Region Proposal Network (RPN) \cite{ren2015faster} generates proposals (\Cref{fig:stage1_penet} \textbf{(c)}) which are used to align three different types of features with corresponding image regions: union features, spatial features, and visual features (\Cref{fig:stage1_penet} \textbf{(d)}, \textbf{(e)}, and \textbf{(f)}) using ROI Align \cite{ren2015faster}. 
Union and spatial features are aggregated for every possible pair of proposals and will serve as the initial edge representation to the relation prediction stage (Stage 2, \Cref{fig:stage2_penet}). A classifier extracts the class labels of each proposal (\Cref{fig:stage2_penet} \textbf{(h)}) and retrieves corresponding linguistic features from pre-trained text embeddings (GloVe \cite{pennington2014glove}). Then, the visual features are combined with the textual features to form the node representation through semantic fusion (\Cref{fig:stage2_penet} \textbf{(i)}, \textbf{(j)}). Then, prototype learning takes place (\Cref{fig:stage2_penet} \textbf{(k)}). Finally, object and predicate labels are decoded to form the final graph representation (\Cref{fig:stage2_penet} \textbf{(l)}, \textbf{(m)}, and \textbf{(n)}). Below, we discuss the main bottlenecks of this architecture for Latency and Object Detection accuracy.

\qquad \textbf{Object Detection Accuracy.} The strategy employed in the two-stage SGG is to decode object labels twice during the relation prediction stage: one time for extracting text features (\Cref{fig:stage2_penet} \textbf{(h)}) and one time to compute the final predictions (\Cref{fig:stage2_penet} \textbf{(m)}). 
The predictions obtained from this last step depend on each model's context learning strategy \cite{tang_learning_2019,zellers_neural_2018,lin_gps-net_2020,tang_unbiased_2020}. This architecture design is (i) redundant as similar operations are performed twice, and (ii) can introduce significant differences in object detection accuracy in contrast to the original Faster-RCNN implementation. \Cref{tab:fastercnn_results} draws a comparison of the performance of the Faster-RCNN model equipped with a ResNeXt-101 feature extractor on the PSG dataset with different relation prediction heads. We can observe that the performance in OD drops significantly, from 1 point wrt Motifs-TDE \cite{zellers_neural_2018,tang_unbiased_2020} to almost 13 points in mAP@50 wrt GPS-NET \cite{lin_gps-net_2020} after the relation prediction stage. These results demonstrate a clear dependency between the two stages of the pipeline, which negatively impacts model performance on the object detection (OD) task. A second consideration given these results is the low performance of Faster-RCNN. Recent state-of-the-art detectors such as YOLOV8/9/10 may be more appropriate for the task. In contrast to Faster-RCNN, the YOLO architecture is not designed for Feature Extraction. This is the main reason why the Faster-RCNN architecture is still widely used for SGG.

\begin{table}[t]
    \begin{center}
    \resizebox{.55\columnwidth}{!}{
        \begin{tabular}{c|c|c}
            \hline
            Backbone & Relation Head & mAP@50 \\
            \hline
            \multirow{6}{*}{Faster-RCNN} & - & 36.4  \\
             & Neural-Motifs \cite{zellers_neural_2018} & 35.9  \\
             & PE-NET \cite{zheng2023prototype} & 35.4\\ 
             & Transformer \cite{tang_unbiased_2020} & 34.6 \\
             & VC-Tree \cite{tang_learning_2019} & 32.6  \\ 
             & GPS-NET \cite{lin_gps-net_2020} & 31.5  \\
             \hline
        \end{tabular}
    }
    \end{center}
    \vspace{0.1in}
    \caption{OD accuracy for two-stage models compared to Faster-RCNN, PSG dataset \cite{yang_panoptic_2022}.}
    \label{tab:fastercnn_results}
\end{table}













\qquad \textbf{Proposal Candidate Selection.} In the TS architecture, Non-Maximum Suppression (NMS) for object proposals (\Cref{fig:stage2_penet} \textbf{(n)}) is performed in stage 2.
To alleviate the lack of NMS in stage 1, a fixed set of the top proposals is traditionally considered valid node candidates for stage 2. The idea here is that by sampling a large number of proposals per image, it is easier to find valid pairs and relations. 
However, this increases the computational complexity of the graph learning. Also, the number of top proposals to select (traditionally 80 or 100) is set arbitrarily. 
Recent work \cite{zellers_neural_2018,tang_unbiased_2020,li_bipartite_2021} computes the final ranking of relations with the following score formula: \(    \theta_{rel} = \theta_{obj} * \theta_{pred} * \theta_{subj},\)
where $\theta_{pred}$ is the confidence score of a predicate, given $<subject, object>$ pair as candidate; and, $\theta_{obj}, \theta_{subj}$ are the respective confidence score of the object detector. 
This formulation assigns greater weight to the confidence scores of the subject and object compared to the predicate. As a result, many low-confidence proposals can be filtered out prior to the relation prediction stage with minimal impact on the model's overall accuracy.

\qquad \textbf{Latency of the PE-NET model.} 
In PE-NET, node prototypes are formed using a combination of linguistic features and visual features (see \Cref{fig:stage2_penet} \textbf{(j)}).
Edge features are computed using the concatenation of the spatial and union features (see \Cref{fig:stage2_penet} \textbf{(g)}). During prototype learning, prototype representations of the subject and object are combined and added to the edge features to form the relation prototype. Finally, a gate mechanism is employed to select relevant features from the multi-modal fusion.
The use of both union and spatial features represents a considerable amount of overload for the model. In fact, another step of ROI Align is performed on the union of subject and object boxes (\Cref{fig:stage1_penet} step \textbf{(d)}). However, visual information is already present in the object and subject features (see \Cref{fig:stage1_penet} \textbf{(f)}), which will be fused with edge features during prototype learning.
In addition, during edge and node feature refinement, all features are traditionally upsampled to a fixed size of 4096 (see \Cref{fig:stage2_penet} \textbf{(g)} and \Cref{fig:stage2_penet} \textbf{(i)}). However, larger feature sizes can significantly augment the latency of the model.
Therefore, we aim to investigate the usage and size of each feature type to find an optimal trade-off between latency and performance.


%% file: sec/4_real_time_SGG.tex
\section{Toward Real-Time SGG}

This work proposes a novel architecture for real-time SGG, as shown in \Cref{fig:stage1_react} and \Cref{fig:stage2_react}. This architecture applies the Decoupled Two-Stage method, replacing the Faster-RCNN detector of PE-NET with a YOLO backbone (cf. \Cref{fig:stage1_react}). The Prototype Learning of PE-NET is improved by removing superfluous components, which results in Efficient Prototype Learning.
This combination of DTS and Efficient Prototype Learning constitutes the REACT model. We also propose Dynamic Candidate Selection (DCS) to further reduce the latency of the proposed model at inference time, as described below.

\qquad \textbf{Decoupled Two-Stage SGG.} To maintain the object detector accuracy during relation prediction training, we propose to freeze the regression and the classification head of the object detector and perform NMS before the relation prediction stage. With this strategy, the object classifier in the relation prediction stage is also removed and class prediction from the object detector is used for the final prediction.
Thus, the relation prediction stage outputs only the pairs and associate predicates, in contrast to also predicting the class labels of objects. 
The loss used during relation prediction:
$\mathcal{L}_{\text{total}} = \mathcal{L}_{\text{rel}} + \mathcal{L}_{\text{obj}}$, then
becomes: $\mathcal{L}_{\text{total}} = \mathcal{L}_{\text{rel}}$, lowering the model complexity and, consequently, its computational load. This approach helps to evaluate the context learning of SGG models independently of the object detector used.
With this architecture, any object detector can be used in place of Faster-RCNN, without impacting stage 2 of the architecture. For instance, one could consider replacing Faster-RCNN with a real-time object detector such as YOLO \cite{jinFastContextualScene2023a}. 
To take advantage of YOLO versions, we leveraged the introduced Decoupled Two-Stage strategy and replaced the ResNeXt-101 backbone with YOLO's CSPNet-based backbones \cite{Jocher_Ultralytics_YOLO_2023,wang2024yolov9} (\Cref{fig:stage1_react} \textbf{(a)}). 
P5-FPN of Faster-RCNN was also replaced by the YOLOV8/9/10 PAN-FPN Necks (\Cref{fig:stage1_react} \textbf{(b)}). Finally, we removed the RPN and detection head of Faster-RCNN and replaced them with the YOLOV8/9/10 Detection Head (\Cref{fig:stage1_react} \textbf{(c)}). Unlike Faster-RCNN, YOLO-based architectures are not designed for feature extraction because they do not implement any ROI Align module. To address this issue, we combined the feature maps extracted from the P3, P4, and P5 layers of YOLO and added one 3x3 Conv block to reduce the feature sizes before ROI Align (\Cref{fig:stage1_react} \textbf{(f)}). The ROI Align module from Faster-RCNN has been modified to align features with corresponding bounding boxes at only three different scales (instead of five in the baseline implementation, \Cref{fig:stage1_react} \textbf{(f)}).

\qquad \textbf{Dynamic Candidate Selection.} Here, we introduce the DCS method to select an optimal number of proposal candidates at inference time. 
During training, the relation prediction stage takes as input a maximum amount of proposals $\theta$ ($\theta=100$). During evaluation, we uniformly sample the number of proposals $k$, starting from 0. Using the generated slope for each metric, the optimal DCS threshold is computed as : 
\(x_{\text{opt}} = \min \left\{ x \mid \left| f'(x) \right| < \epsilon \right\}.\) During inference, $x_{\text{opt}}$ is applied to control the maximum number of proposals to use as input for the relation prediction stage, successfully lowering the complexity of the task.

\qquad \textbf{Efficient Prototype Learning.} In PE-NET, node, and edge features are upsampled 8x and 4x respectively to a fixed size of 4096 by a feed-forward layer (steps \textbf{(g)} and \textbf{(h)} in \Cref{fig:stage2_penet}). We believe this is unnecessary and that the model will perform similarly without upsampling. We then removed this layer and kept the original feature sizes of 256 and 512, respectively. In the latter stage during prototype learning, we removed the associated downsampling layers.
For the PE-NET model, the use of union features for the edge representation seems redundant with the visual features of the subject and object. As a result, we removed the use of union features and dedicated ROI Align step (\Cref{fig:stage1_penet} \textbf{(d)}). By doing so, the edge representation refinement step (\Cref{fig:stage1_penet} \textbf{(g)}) can also be removed and the spatial features fed directly to the context learning (\Cref{fig:stage2_react} \textbf{(i)}). 
To further boost accuracy, we proposed to modify the fusion of subject $s$ and object $o$ features during Prototype Learning of the PE-NET model. We replaced the fusion operation:
$ReLU(s + o) - (s - o)^{2}$
with a single linear layer and removed the subsequent gate operation.


%% file: sec/5_experiments.tex
\section{Experiments}
\begin{table}[t]
    \begin{center}
    \resizebox{\columnwidth}{!}{
        \renewcommand{\arraystretch}{1.3}
        \begin{tabular}{l|l|l|c|c|l|l|l|c}
            \hline
            \textbf{B} & \textbf{D} & \textbf{Relation Head} & \textbf{mR@20/50/100} & \textbf{R@20/50/100} & \textbf{F1@K} & \textbf{mAP\textsuperscript{50}} & \textbf{Latency} & \textbf{Params}  \\ \hline
            \multirow{12}{*}{\rotatebox{90}{\textbf{CSPDarkNet-53}}} & \multirow{12}{*}{\rotatebox{90}{\textbf{YOLOV8m-DTS}}} & REACT (ours) $\dagger$ & \ \underline{18.3} / \ \underline{20.0} / \underline{20.8} & 27.5 / 30.9 / 32.2 & \underline{23.9} & \textbf{53.1} & \underline{23.0} & \underline{43.3M}\\
            & & REACT (ours) & \textbf{18.3} / \textbf{20.1} / \textbf{20.9} & 27.6 / 30.9 / 32.3 & 
            \textbf{23.9} & 53.1 & 32.5 & 43.3M\\  \cline{3-9}
            & & PE-NET $\dagger$ \cite{zheng2023prototype} & 17.1 / 19.0 / 19.8 & \underline{28.0} / \underline{31.3} / 32.9 & 23.2  \scriptsize{$\uparrow$9.1} & 53.1 \scriptsize{$\uparrow$17.6} & 51.7 \space\space \scriptsize{$\downarrow$338} & 187M \\ 
            & & PE-NET \cite{zheng2023prototype} & 17.1 / 19.0 / 19.9 & \textbf{28.0} / \textbf{31.3} / \underline{33.0} & 23.2 \scriptsize{$\uparrow$9.1} & 53.1 \scriptsize{$\uparrow$17.6} & 103.0 \scriptsize{$\downarrow$261} & 187M \\
            & & GPS-NET $\dagger$ \cite{lin_gps-net_2020} & 10.9 / 12.6 / 13.5 & 26.3 / 30.0 / 31.8 & 17.3  \scriptsize{$\uparrow$5.7} & 53.1 \scriptsize{$\uparrow$21.5} & 27.8 \space\space \scriptsize{$\downarrow$286} & 45.3M \\
            & & GPS-NET \cite{lin_gps-net_2020} & 11.1 / 12.7 / 13.7 & 26.6 / 30.1 / 31.9 & 17.6 \scriptsize{$\uparrow$5.9} & 53.1 \scriptsize{$\uparrow$21.5} & 41.1 \space\space \scriptsize{$\downarrow$272} & 45.3M \\
            & & Neural-Motifs $\dagger$ \cite{zellers_neural_2018} & 10.1 / 11.6 / 12.5 & 25.7 / 28.7 / 30.3 & 16.2 \scriptsize{$\uparrow$3.8} & 53.1 \scriptsize{$\uparrow$17.2} & 22.4 \space\space \scriptsize{$\downarrow$296} & 52.9M \\
            & & Neural-Motifs \cite{zellers_neural_2018} & 10.1 / 11.6 / 12.5 & 25.7 / 28.7 / 30.4 & 16.2 \scriptsize{$\uparrow$3.8} & 53.1 \scriptsize{$\uparrow$17.2} & 35.8 \space\space \scriptsize{$\downarrow$282} & 52.9M \\
            & & VCTree $\dagger$ \cite{tang_learning_2019} & 10.7 / 12.2 / 12.9 & 22.1 / 25.3 / 26.8 & 16.1 \scriptsize{$\uparrow$3.7} & 53.1 \scriptsize{$\uparrow$20.5} & 183.2 \scriptsize{$\downarrow$336} & 268.0M \\ 
            & & VCTree \cite{tang_learning_2019} & 10.7 / 12.2 / 13.0 & 22.1 / 25.3 / 26.8 & 16.1 \scriptsize{$\uparrow$3.8} & 53.1 \scriptsize{$\uparrow$20.5} & 239.5 \scriptsize{$\downarrow$280}  & 268.0M \\
            & & Transformer $\dagger$ \cite{tang_unbiased_2020} & 11.3 / 12.4 / 12.8 & 24.1 / 27.0 / 28.3 & 16.6 \scriptsize{$\uparrow$5.5} & 53.1 \scriptsize{$\uparrow$18.5} & \textbf{21.1} \space\space \scriptsize{$\downarrow$275} & 97.9M \\ 
            & & Transformer \cite{tang_unbiased_2020} & 11.3 / 12.4 / 13.2 & 24.1 / 27.1 / 28.9 & 16.9 \scriptsize{$\uparrow$5.7} & 53.1 \scriptsize{$\uparrow$18.5} & 35.1 \space\space \scriptsize{$\downarrow$263} & 97.9M \\ \hline
            \multirow{5}{*}{\rotatebox{90}{\textbf{ResNeXt-101}}} & \multirow{5}{*}{\rotatebox{90}{\textbf{Faster-RCNN}}} & PE-NET \cite{zheng2023prototype} & 10.7 / 11.7 / 12.0 & 16.8 / 18.7 / 19.5 & 14.1  & 35.5 & 390.4 & 426.5M \\
            & & GPS-NET \cite{lin_gps-net_2020} & 8.2 / 8.8 / 9.2 & 15.7 / 18.1 / 19.2 & 11.7 & 31.6 & 313.9 & 391.6M \\ 
            & & Neural-Motifs \cite{zellers_neural_2018} & 8.9 / 9.6 / 9.8 & 16.9 / 18.7 / 19.5 & 12.5 & \underline{35.9} & 318.6 & 369.6M \\
            & & VCTree \cite{tang_learning_2019} & 7.9 / 9.1 / 9.6 & 18.0 / 20.7 / 22.3 & 12.4 & 32.6 & 519.5 & 365.9M \\
            & & Transformer \cite{tang_unbiased_2020} & 7.6 / 8.2 / 8.5 & 16.2 / 17.9 / 18.7 & 11.1 & 34.6 & 296.2 & 328.3M \\ \hline
            \multirow{2}{*}{\rotatebox{90}{\textbf{DETR}}} & \multicolumn{2}{c|}{EGTR \cite{imEGTRExtractingGraph2024}} & 12.0 / 14.5 / 16.6 & 24.9 / 30.3 / \textbf{33.8} & 19.4 & 33.6 & 78.3 & \textbf{42.5M} \\
            & \multicolumn{2}{c|}{RelTR \cite{cong_reltr_2022}} & 7.9 / 10.0 / 11.7 & 12.9 / 17.3 / 20.2 & 12.4 & 25.7 & 59.6 & 63.7M \\
            \hline
        \end{tabular}
    }
    \end{center}
    \vspace{0.2in}
    \caption[Results of experiments with YOLOV8 - full numbers]{Results on the PSG dataset. $\dagger$ represents evaluation with our DCS strategy. Bold and underlined represent the best and second best for each metric. $\downarrow$ and $\uparrow$ represent the improvements of DTS-YOLO against the corresponding baselines.}
    \label{tab:results_2_stage}
\end{table}

\textbf{Datasets.} In SGG, the most widely used dataset is Visual Genome \cite{krishna_visual_2017}. The commonly used split of the data is called VG150 and contains annotations from the top 150 object classes and top 50 predicate classes. However, VG150 suffers from severe biases \cite{yang_panoptic_2022,neau2023fine,hudsonGQANewDataset2019} and inherent issues such as the presence of ambiguous classes (e.g. the classes "people", "men" etc) \cite{yang_panoptic_2022,liang_vrr-vg_2019}. Recently, two new datasets have been proposed with higher-quality annotations: the PSG \cite{yang_panoptic_2022} and IndoorVG \cite{neau2023defense} datasets. Both datasets have been carefully annotated to remove the presence of ambiguous classes or wrong annotations, which makes them a better choice for benchmarking our approach.

\qquad \textbf{Metrics.} 
Following previous work \cite{xu_scene_2017,tang_learning_2019}, we used the Recall@K (R@K) and meanRecall@K (mR@K) metrics to measure the performance of models. For both metrics, we evaluate the recall on the top $K$ ($k=[20,50,100]$) relations predicted, ranked by confidence.
The Recall@K evaluates the overall performance of a model on the selected dataset whereas meanRecall@K evaluates the performance on the average of all predicate classes, which is more significant for long-tail learning such as in the task of SGG. Both metrics are averaged in the F1@K metric as follows:
$F1@K = \frac{2 \times R@K \times mR@K}{R@K + mR@K}$
which efficiently represents the trade-off for the model performance between head and tail predictions. Object Detection (OD) accuracy is measured using standard mAP@50 (mAP\textsuperscript{50}).

\qquad \textbf{Implementation details.} The latest YOLO versions (V8, V9, and V10) can be used interchangeably in our approach. For simplicity here, we will present experiments for the YOLOV8 medium version \cite{Jocher_Ultralytics_YOLO_2023} (YOLOV8m). YOLOV8m has been trained for object detection for 100 epochs on all datasets with a batch size of 16 and image size 640x640.
Relation prediction models have been trained for 20 epochs, batch size 8 with the SGD optimizer, and an initial learning rate of $1e^{-3}$.
For Dynamic Candidate Selection (DCS), we choose $\epsilon = 1e^{-5}$. Inference without DCS is performed with 100 proposals for all models.
In contrast to YOLO, models trained with Faster-RCNN use 600x1000 image sizes.
We fully re-trained seven different relation prediction models on the PSG and IndoorVG datasets: Neural-Motifs \cite{zellers_neural_2018}, VCTree \cite{tang_learning_2019}, Transformer \cite{tang_unbiased_2020}, GPS-Net \cite{lin_gps-net_2020}, PE-NET \cite{zheng2023prototype}, RelTR \cite{cong_reltr_2022}, and EGTR \cite{im2024egtr} following the provided codebases. All models are implemented without any re-weighting \cite{tengStructuredSparseRCNN2022b} or de-biasing strategy \cite{tang_unbiased_2020}. Latency is benchmarked with a batch size of 1\protect\footnotemark.
\footnotetext{Hardware used: \text{11th Gen Intel\textsuperscript{TM}} Core\textsuperscript{TM} i9-11950H @ 2.60GHz x 16, NVIDIA GeForce RTX 3080 Laptop GPU 16GB VRAM, 2 x 16GB 3200 MHz RAM.}

\qquad \textbf{Results.} In \Cref{tab:results_2_stage} and \Cref{tab:results_indoorvg} we display the performance of our DTS architecture in OD accuracy compared to previous work. The F1@K results from \Cref{tab:results_2_stage} are represented as a graph in \Cref{fig:comparison_intro}. In the text, we refer to improvements as mean absolute percentage.
\begin{table}[t]
\begin{minipage}{.49\textwidth}
\begin{center}
    \resizebox{\columnwidth}{!}{
    \renewcommand{\arraystretch}{1.2}
    \begin{tabular}{l|l|c|c|c|c}
        \hline
            \textbf{B} & \textbf{Relation Head} & \textbf{mR@K} & \textbf{R@K} & \textbf{F1@K} & \textbf{mAP\textsuperscript{50}} \\ \hline
            \multirow{7}{*}{\rotatebox{90}{\textbf{DTS}}} 
            & REACT (ours) & \textbf{18.0} & \textbf{30.9} & \textbf{22.8} & \textbf{37.2}\\ 
            & PE-NET & \underline{16.0} & 29.4 & \underline{20.7} & 37.2\\ 
            & GPS-NET & 9.9 & 29.2 & 14.8 & 37.2 \\ 
            & Neural-Motifs & 11.1 & \underline{30.5} & 16.2 & 37.2\\
            & VCTree & 14.9 & 18.9 & 16.7 & 37.2\\ 
            & Transformer & 12.3 & 28.4 & 17.2 & 37.2\\ \hline
            \multirow{5}{*}{\rotatebox{90}{\textbf{TS}}} & PE-NET & 13.8 & 27.1 & 18.1 & 25.2 \\ 
            & GPS-NET & 6.7 & 11.9 & 8.6 & 17.4 \\ 
            & Neural-Motifs & 9.8 & 15.2 & 11.9 & \underline{26.2}\\ 
            & VCTree & 11.0 & 14.7 & 12.6 & 25.5 \\ 
            & Transformer & 9.9 & 16.8 & 12.5 & 24.2 \\ \hline
            \multirow{2}{*}{\rotatebox{90}{\textbf{OS}}} & EGTR \cite{imEGTRExtractingGraph2024} & 7.1 & 10.6 & 8.5 & 14.7\\ 
            & RelTR \cite{cong_reltr_2022} & 7.7 & 11.4 & 9.2 & 14.1\\ \hline
        \end{tabular}
    }
    \end{center}
    \vspace{0.2in}
    \caption{Results on IndoorVG. mR@K and R@K are average for $k=[20,50,100]$.}
    \label{tab:results_indoorvg}
\end{minipage}
\hfill
\begin{minipage}{.49\textwidth}
\begin{center}
    \resizebox{\columnwidth}{!}{
        \renewcommand{\arraystretch}{1.2}
        \begin{tabular}{l|l|c|c|c|c}
        \hline
            \textbf{B} & \textbf{Model} & \textbf{mR@K} & \textbf{R@K} & \textbf{F1@K} & \textbf{mAP\textsuperscript{50}} \\ \hline
            \multirow{1}{*}{\textbf{DTS}}
            & REACT & 12.9 & 27.4 & 17.6 & \textbf{31.8} \\ \hline \multirow{4}{*}{\textbf{TS}} & PE-NET & 13.3 & 32.6 & \underline{18.9} & 29.2 \\ 
            & SQUAT \cite{jung2023devil} &  \textbf{15.3} & 26.7 & \textbf{19.4} & - \\
            & BGNN \cite{li_bipartite_2021} & 11.6 &  \underline{33.4} & 17.3 & 29.0 \\
            & VCTree & 7.2 & \textbf{33.9} & 11.8 & 28.1 \\ 
            \hline
            \multirow{4}{*}{\textbf{OS}} & EGTR \cite{imEGTRExtractingGraph2024} & 7.8 & 29.3 & 12.4 & \underline{30.8} \\ 
            & SGTR \cite{li_sgtr_2022} & \underline{13.6} & 26.5 & 18.0 & 25.4 \\ 
            & SS-RCNN \cite{tengStructuredSparseRCNN2022b} & 9.5 & 36.0 & 15.0 & 23.8 \\ 
            & RelTR \cite{cong_reltr_2022} & 8.0 & 24.3 & 12.0 & 26.4 \\ \hline
        \end{tabular}
    }
    \end{center}
    \vspace{0.2in}
    \caption{REACT compared to previous work on VG150 \cite{xu_scene_2017,krishna_visual_2017} (reported results). mR@K and R@K are the average for $k=[50,100]$ as some approaches do not report $k=20$.}
    \label{tab:results_vg150}
\end{minipage}
\end{table}
By using YOLOV8, we improved the mAP by 54.37\% on the PSG dataset (\Cref{tab:results_2_stage}) and 69.92\% on IndoorVG (\Cref{tab:results_indoorvg}), compared to TS approaches with Faster-RCNN. Compared to OS approaches (EGTR \cite{im2024egtr} and RelTR \cite{cong_reltr_2022}), mAP is improved by 120\% on average, which demonstrates the inefficiency of such approaches for the OD task.
For a fair comparison, we also present the performance of the REACT model on the VG150 dataset, see \Cref{tab:results_vg150}. The performance of REACT in mAP is better by one point than the best approach EGTR. We believe that the low quality of the bounding box annotations and the ambiguous classes in VG150 are confusing the YOLOV8 detector, which leads to only a small improvement against Faster-RCNN.
On the PSG dataset, we experienced an average improvement of 58.47\% in F1@k by using our DTS method compared to the same approaches with Faster-RCNN. In addition, our REACT model outperforms the PE-NET model with DTS by a small margin (23.9 versus 23.2 in F1@K). We observe that the gain in F1@k is not very strongly correlated to the gain in mAP between the Two-Stage and DTS approaches, as we measured a correlation coefficient of 0.75 between the two.
On the IndoorVG dataset, we observe an improvement of 38.76\% in F1@K with the DTS approach. Our REACT model also outperforms PE-NET (DTS) by a strong margin (22.8 versus 20.7 in F1@K).
We observe that REACT outperforms some Two-Stage approaches (Motifs, VCTree, and BGNN) for RelPred on VG150 (see \Cref{tab:results_vg150}) but is not competitive with more recent approaches (PE-NET and SQUAT). We believe this is due to noisy feature representation. Because REACT has fewer parameters, it is more sensitive to noisy data.
Regarding latency, we demonstrated an improvement of 84.99\% using YOLOV8 instead of Faster-RCNN, which is a considerable gap. Regarding the number of parameters, the DTS approach uses on average 77.47\% less parameters than the baseline.
Our introduced REACT model is the fastest one, with a latency of 32.5ms. 
By applying our DCS strategy, we gain an average of 66.5\% latency on all models (compared to the YOLOV8m-DTS versions) with a relative average loss of 1\% in F1@K.
The EGTR model stays the best overall in terms of parameter count by a slow margin (42.5M versus 43.3M for REACT).
By comparing REACT to the fastest model benchmarked, RelTR \cite{cong_reltr_2022}, REACT is 2.7x faster with $+10$ points on average in F1@K and $+24.9$ points on average in mAP\textsuperscript{50} for the 3 datasets.
By taking into account OD accuracy, RelPred accuracy, and latency, our REACT model is the best compromise. Even for the VG150 dataset, REACT is a better choice because no other two-stage approaches are close to real-time inference. On the other hand, when evaluated fairly, one-stage approaches are clearly behind in terms of mAP and F1@K on all datasets, and are then not a good choice in comparison to REACT.

%% file: sec/6_ablation_studies.tex


\qquad \textbf{Latency and Number of Proposals.} As explained before, the second stage of SGG takes as input a fixed number of $n$ proposals as node candidates. There could be a maximum of $n * (n-1)$ possible pairs in the graph, thus when doing matching the computational complexity is supposed to scale accordingly. To demonstrate this hypothesis, we evaluated the performance of the REACT model in both latency and accuracy for different numbers of input proposals $k=\{10,150\}$ with DTS-YOLOV8 and with classical Faster-RCNN. For all experiments, we ranked proposals by confidence of the backbone and selected the top $k$.
\begin{figure}[t]
    \centering
    \begin{subfigure}{.49\textwidth}
    \centering
        \includegraphics[width=\columnwidth]{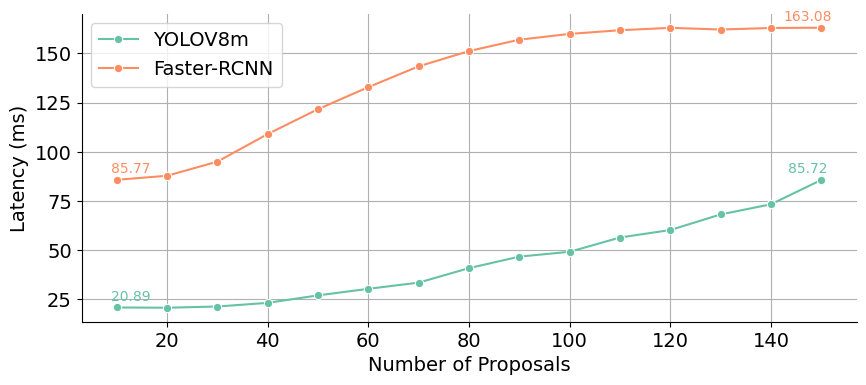}
    \end{subfigure}
    \begin{subfigure}{.49\textwidth}
    \centering
        \includegraphics[width=\columnwidth]{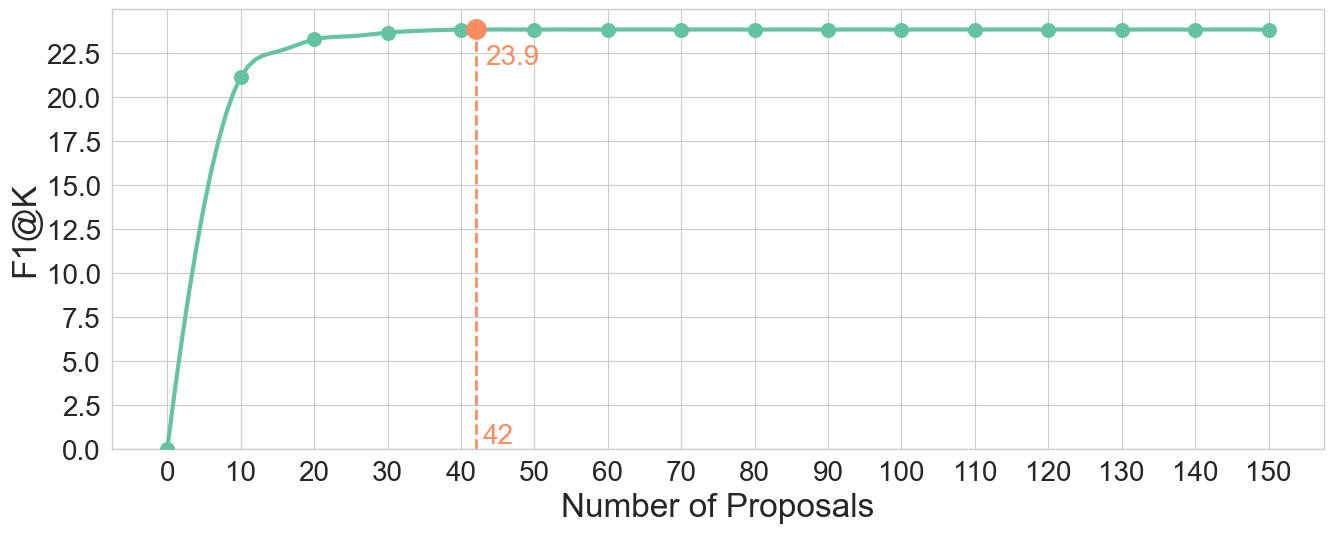}
    \end{subfigure}
    \caption{Left: Latency for the REACT model using a different number of proposals per image, with batch size 1. Right: Average F1@k for the REACT model with DTS-YOLO.}
    \label{fig:num_proposals}
\end{figure}
The results of these experiments are shown in \cref{fig:num_proposals} (top), where we observe a plateau phenomenon for REACT with Faster-RCNN above 100 proposals. This behavior does not appear when using DTS+YOLOV8. By using 150 proposals with DTS+YOLOV8, the latency was similar to 10 proposals with Faster-RCNN. The evaluation of the REACT model under different sampling conditions with our DCS strategy is represented in \cref{fig:num_proposals} (bottom). In this figure, we can see that the optimal value was found at 23.89 for $k=42$, just before that the performance starts to be saturated.

\qquad \textbf{Features Fusion.} During context learning of the REACT model, and more generally in a lot of approaches in SGG \cite{tang_unbiased_2020,zellers_neural_2018,lin_gps-net_2020,zheng2023prototype}, different feature sources are used to model the edge representation. First, union and spatial features are combined and then merged with the addition of the subject and object node features. The node features are created using visual and textual features obtained for each proposal. 
In our REACT model, we removed the dependency on the union features (see \Cref{fig:stage2_react} \textbf{(i)}). In the following, we justify our choice by analyzing the impact of each feature type on learning.
The results from these experiments are displayed in \Cref{tab:features}. The baseline REACT model is the second row. With union features (first row) we observe a significant decrease in F1@K as well as latency. We hypothesize that union features are too noisy and confuse the model. This is consistent with previous work which highlights the importance of spatial features for the task \cite{jin_fast_2023}.
In the last two rows, we removed all dependencies on contextual features (union and spatial) and kept only the concatenation of the subject and object node as initial edge features. We observe only a slight decrease by keeping only text features (last row). In fact, the model can retain up to 71\% of the optimal performance by learning only from the textual GLoVe features. This suggests a trade off between text and visual features in the task.

\begin{table}[t]
    \begin{center}
    \resizebox{.75\columnwidth}{!}{
        \begin{tabular}{c|c|c|c|c|c|c|c|c}
            \bf T & \bf V & \bf S & \bf U & \bf mR@K  & \bf R@K  & \bf F1@K & \bf Latency & \bf Params \\
            \hline
            \cmark & \cmark & \cmark & \cmark & \underline{14.6} & \textbf{30.8} & \underline{19.8} & 36.0 & 42.7M \\
            \cmark & \cmark & \cmark & \xmark & \textbf{19.7} & \underline{30.2} & \textbf{23.9} & 32.5 & 42.7M \\
            \cmark & \cmark & \xmark & \cmark & 14.2 & 29.9 & 19.3 & 27.3 & 42.7M \\
            \cmark & \cmark & \xmark & \xmark & 14.5 & 30.1 & 19.6 & \underline{18.5} &  \underline{36.6M} \\
            \cmark & \xmark & \xmark & \xmark & 13.5 & 29.1 & 18.4 & \textbf{16.2} &  \textbf{35.2M} \\
        \end{tabular}
    }
    \end{center}
    \vspace{0.2in}
    \caption[Ablation study: features selection]{Feature sources experiments. V and T are visual and textual features for $<subject, object>$, and S and U are Spatial and Union features, respectively.}
    \label{tab:features}
\end{table}



%% file: sec/8_conclusion.tex
\section{Conclusion}

In this work, we proposed a novel implementation of a real-time Scene Graph Generation model that attains competitive performance with a latency of 23ms. We proposed a new architecture, the Decoupled Two-Stage SGG (DTS), aiming at taking advantage of real-time object detectors for SGG. We also proposed a new inference method, Dynamic Candidate Selection (DCS) to further reduce latency by 66.5\% on average without compromising performance. Finally, we proposed a new model, REACT, which combines the DTS architecture with efficient prototype learning for lower latency and improved accuracy.
Our main finding is that the performance and latency of SGG models are heavily correlated with the quality and quantity of object proposals generated by the object detector. However, regarding both overall performance and latency, a trade-off can be found by using our DCS strategy.
Another finding is the over-reliance of models on the text features for relation prediction.
Future work will consider implementing the REACT model in constraint settings such as embodied agent navigation or reasoning. Due to its small size, REACT can be embedded onboard robotic platforms and provide reliable and fast predictions that can foster reasoning.

\section*{Acknowledgments}

\noindent This work has been supported by the Brittany Region and the Knut and Alice Wallenberg foundation and the Wallenberg AI, Autonomous Systems and Software Program (WASP).